\documentclass[sigconf]{acmart}

\AtBeginDocument{%
  }

\setcopyright{acmlicensed}
\copyrightyear{2018}
\acmYear{2018}
\acmDOI{XXXXXXX.XXXXXXX}
\acmConference[Conference acronym 'XX]{Make sure to enter the correct
  conference title from your rights confirmation email}{June 03--05,
  2018}{Woodstock, NY}
\acmISBN{978-1-4503-XXXX-X/2018/06}

\usepackage{graphicx}
\usepackage{amsmath}
\usepackage{amsthm}
\usepackage{booktabs}
\usepackage{graphicx}
\usepackage{algorithm}
\usepackage{svg}
\usepackage{tabularx}
\usepackage{multirow}
\usepackage{enumitem}
\usepackage{subcaption}
\usepackage{algpseudocode}
\usepackage{graphicx} % For centering and better alignment
\usepackage{adjustbox} % Optional, for scaling tables
\usepackage[switch]{lineno}
\usepackage{hyperref}
\newcommand{\partitle}[1]{\smallskip \noindent \textbf{#1.}}

\begin{document}

\title{Node-level Contrastive Unlearning on Graph Neural Networks}

\author{Hong kyu Lee}
\email{hong.kyu.lee@emory.edu}
\affiliation{%
  \institution{Emory University}
  \city{Atlanta}
  \state{Georgia}
  \country{USA}
}

\author{Qiuchen Zhang}
\email{qzhan84@emory.edu}
\affiliation{%
  \institution{Emory University}
  \city{Atlanta}
  \state{Georgia}
  \country{USA}
}

\author{Carl Yang}
\email{j.carlyang@emory.edu}
\affiliation{%
  \institution{Emory University}
  \city{Atlanta}
  \state{Georgia}
  \country{USA}
}

\author{Li Xiong}
\email{lxiong@emory.edu}
\affiliation{%
  \institution{Emory University}
  \city{Atlanta}
  \state{Georgia}
  \country{USA}
}



\begin{abstract}
Graph unlearning aims to remove a subset of graph entities (i.e. nodes and edges) from a graph neural network (GNN) trained on the graph. Unlike machine unlearning for models trained on Euclidean-structured data, effectively unlearning a model trained on non-Euclidean-structured data, such as graphs, is challenging because graph entities exhibit mutual dependencies. Existing works utilize graph partitioning, influence function, or additional layers to achieve graph unlearning. However, none of them can achieve high scalability and effectiveness without additional constraints.
  In this paper, we achieve more effective graph unlearning by utilizing the embedding space.
  The primary training objective of a GNN is to generate proper embeddings for each node that encapsulates both structural information and node feature representations. Thus, directly optimizing the embedding space can effectively remove the target nodes’ information from the model.
  Based on this intuition, we propose node-level contrastive unlearning (Node-CUL). It removes the influence of the target nodes (unlearning nodes) by contrasting the embeddings of remaining nodes and neighbors of unlearning nodes.
  Through iterative updates, the embeddings of unlearning nodes gradually become similar to those of unseen nodes, effectively removing the learned information without directly incorporating unseen data.
  In addition, 
  we introduce a neighborhood reconstruction method that optimizes the embeddings of the neighbors in order to remove influence of unlearning nodes to maintain the utility of the GNN model.
  Experiments on various graph data and models show that our Node-CUL achieves the best unlearn efficacy and enhanced model utility with requiring comparable computing resources with existing frameworks. 
  
\end{abstract}




\received{20 February 2007}
\received[revised]{12 March 2009}
\received[accepted]{5 June 2009}

\maketitle

\section{Introduction}

With recent regulations on data privacy such as General Data Protection Regulation (GDPR)~\cite{mantelero_eu_2013} and California Consumer Privacy Act (CCPA)~\cite{pardau2018california}, machine unlearning has gained significant attention from various research communities. Specifically, ``the right to be forgotten'' grants individuals the right to complete removal of their data. This includes the  removal of its influence on the parameters of any machine learning models trained with the data. Accordingly, machine unlearning aims to selectively remove a subset of training data from machine learning models~\cite{cao2015towards}. A successfully unlearned model should achieve three objectives: 1)  unlearning effectiveness, evaluated by how thoroughly the unlearning algorithm removes the target samples, 2) high model utility, assessed by the performance on the original task, and 3) high efficiency, measured by the time and resources needed for the unlearning algorithm.

Graph neural networks (GNNs) have achieved remarkable success in analyzing complex graph data in various research fields, from social media analysis~\cite{Borisyuk2024lignn} to financial fraud detection~\cite{Motie2024financial}. GNNs are designed to effectively learn node representations that can be utilized for downstream tasks such as node classification or link prediction. Along with attempts to ensure privacy on GNN models~\cite{zhang2024dpar, dai2024comprehensive}, there have been efforts to conduct machine unlearning on GNN models, or graph unlearning. Primarily, graph unlearning refers to removing the influence of a set of target graph entities (node features or edges or nodes) from the model. Existing works on graph unlearning can be divided into two types: SISA~\cite{bourtoule2021machine} based and loss-based. In SISA, the training data is divided into multiple shards, and a model is trained separately for each shard. Upon an unlearning request of a sample, SISA identifies which shard has the target sample and retrains the corresponding model with the shard excluding the sample. Various works employed SISA for GNN unlearning~\cite{Chen2022graph,wang2023inductive,Zhang2024graph} by leveraging graph partitioning strategies that effectively partition the graph with minimal information loss.

The loss-based GNN unlearning conducts gradient updates to the model to remove the influence of the target graph entities~\cite{wu2023certified,chien2022certified,zhang2023graph, cheng2023gnndelete,wu2023gif}. Specifically, they introduce unlearning loss to quantify the residual influence of target entities in the model, and conduct parameter updates accordingly to remove their influence. Graph unlearning based on certified unlearning~\cite{wu2023certified, chien2022certified} mathematically guarantees unlearning effectiveness. These frameworks usually leverage convexity of linear GNNs for their analysis. GNNDelete~\cite{cheng2023gnndelete} and GIF~\cite{wu2023gif} are the state-of-the-art frameworks for unlearning non-linear GNNs. GNNDelete proposed deleted edge consistency and neighborhood influence as unlearning losses. The former ensures unlearning efficacy by guiding the model to make random predictions for the target entities, and the latter reduces the impacts of unlearning on subgraphs that are in proximity to the entities. GIF conducts a one-shot gradient update using the modified influence function that can reverse the influence of a target component across the model~\cite{wu2023gif}. 

Due to the unique nature of GNNs, existing works on loss-based GNN unlearning come with several limitations. GNNs learn a node representation from its own embeddings and its neighbors' embeddings through neighborhood aggregation. Thus, when unlearning, it is crucial to reverse the impact of neighborhood aggregations by (1) disconnecting the neighborhood aggregation between the target entities and their neighbors, and (2) Decrease the dependency of all k-hop neighbors on the target entities. GNNDelete solves this problem by inserting an unlearning layer - a feed-forward layer - between each graph convolution layer. During unlearning, only unlearning layers are optimized based on the unlearning loss. During inference, unlearning layers are activated only when processing the embeddings of unlearning nodes. While this effectively reduces neighborhood aggregation, it requires keeping track of every unlearning entity for inference even after unlearning which is not practical. GIF utilizes the influence function to quantify the impact of neighborhood aggregation on target entities and their impacts on their neighbors. Their theoretical analysis only applies to convex or one-layer GNNs. While they empirically evaluated the efficacy of their framework for non-linear GNNs, a comprehensive analysis of the influence function remains an open challenge.

\begin{figure}
    \centering
    \includegraphics[width=0.7\linewidth]{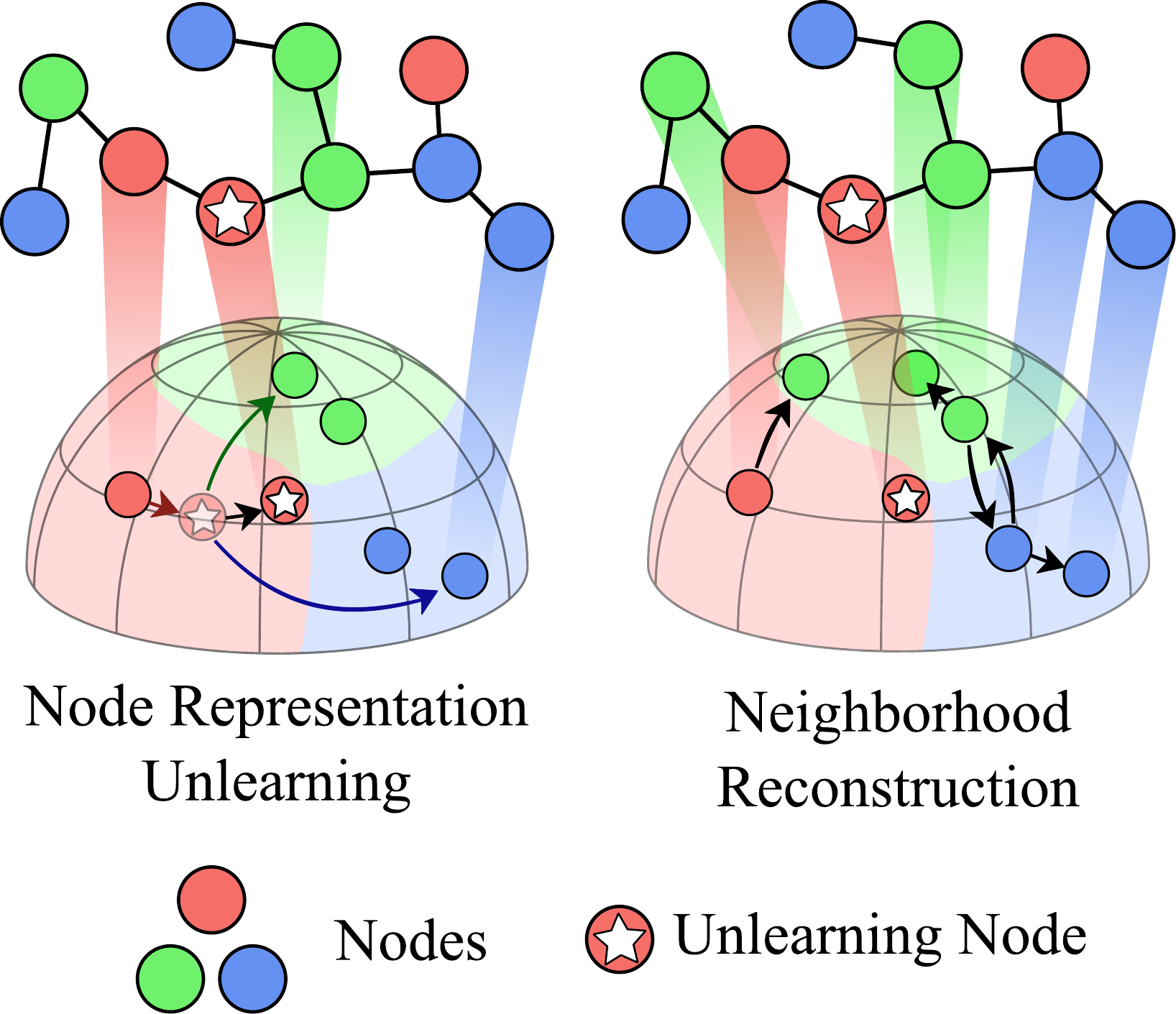}
    \caption{Conceptual visualization of node-level graph contrastive unlearning (Node-CUL)}
    \label{fig:node_level_graph_contrasitve}
    \vspace{-1em}
\end{figure}

\partitle{Our contribution} To address the aforementioned limitations, we propose Node-CUL, a novel node-level contrastive graph unlearning framework that directly optimizes the embedding space via a contrastive approach. It aims to unlearn a set of nodes (unlearning nodes) from a GNN model (node-classification model), which is generally  more challenging than unlearning a set of edges. %Initially, the unlearning nodes are part of the training nodes of the model; however, at the end of unlearning, the model must perceive them as if they are unseen nodes. 
%Also, while each unlearning node itself is considered as an unseen node, the model should perceive some of its neighbor nodes correctly if they were part of the training nodes. 
An unlearned model should perceive the unlearning nodes as if they are unseen nodes while preserving the information of their neighboring nodes accurately.
We reformulate the node-level unlearning problem in the embedding space. As GNNs aim to learn node embeddings that represent both the node and structural information, direct optimization on the embeddings space can effectively remove the node's influence from the model. We adjust the embedding so that those of unlearning nodes become indistinguishable from those of test nodes (unseen nodes), without directly using the test nodes.
%In summary, we modify parameters of GNN models so that 
In this way, the unlearned model behaves similarly for the unlearning nodes and unseen nodes.

To achieve this, we build upon the concept of contrastive unlearning from classification models trained on Euclidean-structured data~\cite{zhang2024contrastive} but with new designs to tackle the unique challenges posed by neighborhood aggregation for graph unlearning. It consists of two components: node representation unlearning and neighborhood reconstruction. The first focuses on effective unlearning of the unlearning nodes, and the second aims to  reverse the neighborhood aggregation from the neighbors to maintain the model utility.
%It proposed unlearning image classification models by directly optimizing the embeddings of them. The framework obtains embeddings of an unlearning sample and remaining samples. Then it updates embeddings of the unlearning sample by (1) pulling them towards the embeddings of remaining samples with different classes and (2) pushing them away from the embeddings of remaining samples with the same class. To this end, embeddings of the unlearning sample are pushed near to the decision boundary of the model, where embeddings of unseen samples are usually located. Accordingly, the model perceives the unlearning samples not as training samples but as if they were unseen. The framework is more effective than naive unlearning frameworks such as gradient ascent or fine-tuning because its impact on the decision boundary is small, preserving the utility of the unlearned model.
%Directly applying the contrastive unlearning for node-level graph unlearning is ineffective as the original framework does not consider the GNNs' neighborhood aggregation. Thus, we advance the idea of contrastive unlearning in the context of node-level graph unlearning. 
We obtain embeddings of an unlearning node, embeddings of its neighbors, and remaining nodes (rest of the training nodes). We perform node representation unlearning by (1) pulling the unlearning node's embedding towards embeddings of remaining nodes with the different class, and (2) pushing the unlearning node's embedding away from embeddings of its one-hop neighbors with the same class. This approach is grounded in two key intuitions. Pulling helps disassociate the unlearning node's class from the model, steering the model to perceive the node as an unseen node. Pushing complements this by disconnecting the unlearning node from its neighbors, encouraging the model to regard unlearning nodes and their neighbors as unrelated. 

One impact of adjusting the embeddings for unlearning is that the neighboring nodes' embeddings will be affected which can affect the model utility. To maintain model utility, we propose neighborhood reconstruction that maintains the model utility by erasing influences of the unlearning nodes on their neighbors.  %further address the unique challenge of graph unlearning. It adjusts the embeddings of neighbors of the unlearning node so that they can be less associated with the unlearning node, and more associated with their neighbors (excluding the unlearning node) and their own class. To this end, embeddings of neighbors are less affected by the embeddings of the unlearning node. 
Specifically, we optimize embeddings of the $k$-hop neighbors of unlearning nodes by pulling them closer to their remaining neighbors (excluding the unlearning nodes) to encourage their association with the unaffected remaining neighbors. 

Figure~\ref{fig:node_level_graph_contrasitve} illustrates a conceptual depiction of  Node-CUL. The color of each node represents its class, the sphere shows the representation space, and the color of the surface of the sphere shows the corresponding decision boundaries. From a node prediction model, neighboring nodes and nodes with the same class have relatively similar representations, while nodes with different classes have dissimilar embeddings. This observation is supported by several studies~\cite{das_separability_2019,keriven2022not, dong2024smoothgnn}. The left shows our node representation unlearning where %figure is a visualization of our unlearning framework. We leverage contrastive unlearning~\cite{zhang2024contrastive} to push 
embeddings of the unlearning node are pushed away from its immediate neighbors with the same class and pulled towards the remaining nodes with different classes. This pushes node embeddings of the unlearning nodes closer to the decision boundary, where embeddings of unseen nodes usually are. The right illustrates our neighborhood reconstruction which adjusts embeddings of neighbors of unlearning nodes by pushing them towards embeddings of their remaining neighbors except the unlearning node, eliminate influences of unlearning nodes from their embeddings for model utility. %This approach aims to prevent the model from allowing the unlearning node to influence the predictions of its neighboring nodes. 

In summary, our contributions are as follows.

\begin{enumerate}
    \item We propose node-level graph contrastive unlearning (Node-CUL) for graph unlearning in the representation space. It is a novel, model-agnostic unlearning framework that utilizes node-level contrastive loss. We utilize node embeddings of unlearning nodes, neighboring nodes, and remaining nodes to effectively remove the influence of the unlearning node.

    \item Our framework consists of two components that complement each other to achieve effective unlearning and high model utility: node representation unlearning and neighborhood reconstruction. Node representation unlearning modifies the embeddings of unlearning nodes so that they behave similarly to unseen test nodes.  %To increase unlearn efficacy and performance, we propose neighborhood reconstruction. It steers the model to prevent the influence of the unlearning node during the prediction of its neighboring nodes. This 
    Neighborhood reconstruction aims to maintain model utility by modifying the embedding of all neighbors of the unlearning nodes to minimize their the reliance on the unlearning nodes. We also incorporate cross entropy loss on all remaining nodes and neighboring nodes to maintain the model utility and design an explicit termination condition to allow the unlearning process to stop properly.  %removing the influence of unlearning nodes from the neighbors' perspective and further enhances the model's prediction performance on the neighbors.

    \item We conduct comprehensive experiments to compare Node-CUL with state-of-the-art loss-based graph unlearning frameworks to demonstrate the effectiveness and versatility of our framework. We also conduct a membership inference attack~\cite{carlini2022membership} to validate and compare the effectiveness of unlearning.
\end{enumerate}

\section{Related Works}

Since machine unlearning was introduced~\cite{cao2015towards}, various  unlearning algorithms  have been proposed~\cite{Golatkar2020eternal,guo2020certified,bourtoule2021machine,izzo2021approximate,Zhang2024towards,tarun2024fast,zhang2024contrastive}. Typically, they can be categorized into exact and approximate unlearning. Exact unlearning completely removes the knowledge of the unlearning samples. SISA is an exact unlearning framework with fast retraining of its sub-models~\cite{bourtoule2021machine}. SISA splits the training dataset into shards and trains a model with each shard. Upon an unlearning request, SISA retrains the model whose shard includes the unlearning samples. Although it completely removes the influence of the unlearning sample, it has limited scalability. For approximate unlearning, Fisher is an unlearning framework that reduces the parametric distance between the model and the model trained without the unlearning samples~\cite{Golatkar2020eternal}. Gradient ascent and fine-tuning are compared as baselines. Certified unlearning~\cite{guo2020certified} %conducts unlearning via a one-shot update to the model. The authors 
conducts a noisy second-order update using the influence function~\cite{ling1984residuals}. Certified unlearning provides a mathematical guarantee of unlearning a linear model by bounding the maximum error. More recent works attempted to extend the analysis to deep models~\cite{Zhang2024towards}. While these works primarily focus on unlearning a sample, several works explore special unlearning scenarios such as unlearning an entire class~\cite{kurmanji2023towards, tarun2024fast} or unlearning several features~\cite{Warnecke2023machine}.

There are increasing efforts on unlearning GNNs~\cite{Chen2022graph, dong2024idea, wu2023certified, wu2023gif, pan2023unlearning, Li_Zhao_Wu_Zhang_Li_Wang_2024, zhang2023graph}. Several graph unlearning frameworks employ SISA~\cite{Chen2022graph,wang2023inductive,Zhang2024graph}. These works adopted a graph partitioning strategy to partition shards with small information loss. Wang \textit{et al.} explored graph unlearning in the inductive setting and demonstrated the effectiveness of the SISA based frameworks. Some of them utilize a graph attention mechanism to boost the model utility~\cite{Zhang2024graph}. Unlearning tasks for these frameworks are primarily node unlearning from a node classification model. A fundamental limitation of these frameworks is the scalability as they need re-training of models. 

Among approximate unlearning, certified unlearning has been utilized for graph unlearning~\cite{chien2022certified, wu2023certified, pan2023unlearning,dong2024idea}. Primarily certified unlearning focuses on linear-GNNs for their frameworks. This is because linear-GNNs meets the requirements of certified unlearning as loss convexity and model linearity are crucial for mathematical analysis. %Chien \textit{et} al. first proposed certified unlearning on graphs~
\cite{chien2022certified} 
extends the original certified unlearning to node feature, edge, and node unlearning from GNN models and ~\cite{wu2023certified} improved edge unlearning. While these frameworks update the entire parameters, ~\cite{pan2023unlearning} utilized graph scattering transforms for processing graph data and certified unlearning on the linear prediction layers. More recently, ~\cite{dong2024idea} enhanced analysis and demonstrated obtaining tighter error bounds is feasible. A critical limitation of these frameworks is that they do not scale to non-linear GNN models.

Aside from certified unlearning, GNNDelete first proposed approximate unlearning on non-linear GNN models~\cite{cheng2023gnndelete} by optimizing unlearning loss of deleted edge consistency and neighborhood influence. More recent work~\cite{yang2023when} boosted performance by contrasting two losses, expediting the optimization process. GIF captures how a target node influences the prediction of neighboring nodes via the influence function~\cite{wu2023gif}. Some frameworks utilize Kullback-Leibler divergence for graph unlearning~\cite{Li_Zhao_Wu_Zhang_Li_Wang_2024,zhang2023graph}. These frameworks utilize reference models for guiding the target model to maintain model performance. More recently, SUMMIT~\cite{Zhang2024forgetting} was proposed as an edge unlearning framework which considers graph structures effectively to reduce performance loss. Each framework has with its own limitations: GNNDelete requires manual recording of target entities, GIF lacks comprehensive analysis on non-linear models, and frameworks based on knowledge distillation demand substantial computing resources. We choose GNNDelete and GIF as baselines to compare as these frameworks exhibit outstanding performance.

Unlearning can be also utilized for enhancing the robustness of GNNs. 
A number of works demonstrated unlearning is effective for mitigating backdoor attack samples~\cite{wang2024made,zhang2024no} and to remove poisoned samples from GNNs and increase performance~\cite{wu2024graphmu}.

\section{Problem definition}
\label{sec:prob_def}
\partitle{Graph Neural Networks (GNNs)}
Let $G = \{V, E, X, Y\}$ be a graph where $V$ is a set of $n=\lvert N \rvert$ nodes, $E$ is a set of edges, $X = \{x_0, \cdots, x_{n-1} \}$ is a set of node features and $Y = \{y_0, \cdots y_{n-1}\}$ is a set of corresponding labels. We denote $f = f_\mathcal{H}\left(f_{\mathcal{E}}\left(\cdot\right)\right)$ where $f_{\mathcal{E}}\left(\cdot\right)$ consists of GNN layers and produces node embeddings, and $f_\mathcal{H}\left(\cdot\right)$ is a prediction head. A layer $f^l_{\mathcal{E}}$ of a GNN model $f$ receives previous embeddings  $h^{l-1}$ and $G$, and provides new node embeddings $h^l$. 
The process of a GNN layer is twofold: aggregation passing and update. For a node $u$, $f^l_{\mathcal{E}}$ first aggregates messages from every neighbor of $u$. Then it obatins new embedding $h^l_u$ via combining $p^l_u$ with its previous embeddings. Formally they can be described as follows:
\begin{align}
    &p^l_u = \textbf{Agg}\left(h^{l-1}_u, h^{l-1}_v, E_{uv} \vert \forall v \in \mathcal{N}^1_u\right) \\
    &h^l_u = \textbf{Upd}\left(h^{l-1}_u, p^l_u\right)
\end{align}
Where $\textbf{Agg}$ and $\textbf{Upd}$ are aggregate and update function and $\mathcal{N}^1_u$ is a one-hop neighbor of $u$. The aggregation and update functions are implemented differently across different GNN architectures.

\partitle{Transductive node-level unlearning}
We primarily focus on unlearning a node classification GNN model trained with a transductive graph, where test nodes are accessible but not optimized during training. Specifically, let $f$ be a node classification model  trained on training nodes $V_{tr}$. Let $V_{ts} = V \backslash V_{tr}$ be the test nodes.
Let $V_{u} \subset V_{tr}$ be a set of nodes to unlearn, and $V_{r} = V_{tr} \backslash V_{u}$ be a set of remaining nodes. With the original model $f$, $V_{u}$ and $V_{r}$, an unlearned model $f^\prime$ should achieve following.
\begin{align}
\mathbf{Acc}\left(f^\prime\left(V_{u}\right), Y_{u}\right) \approx \mathbf{Acc}\left(f^\prime\left(V_{ts}\right), Y_{ts}\right)
\label{eq:prob1}
\end{align}
\begin{align}
\mathbf{Acc}\left(f^\prime\left(V_{ts}\right), Y_{ts}\right) \approx \mathbf{Acc}\left(f\left(V_{ts}\right), Y_{ts}\right)
\label{eq:prob2}
\end{align}
where $\mathbf{Acc}$ is a readout function to measure prediction accuracy of $f$ and $f^\prime$. Equation~\ref{eq:prob1} ensures  effective unlearning, as it requires $f^\prime$ to exhibit similar accuracy on test nodes and unlearning nodes, mirroring the performance of a retrained model that excludes the unlearning nodes. %It requires the model to behave similarly when unlearning nodes and test nodes (unseen nodes) are given. 
Equation~\ref{eq:prob2} ensures model utility, as it requires the unlearned model to maintain similar accuracy  to the original model on test nodes. We validate the problem definition in Section~\ref{sec:exp_model_utility} by empirically demonstrating that the fully re-trained model (gold standard) satisfies Equations~\ref{eq:prob1} and~\ref{eq:prob2}.

\section{Node-level Contrastive Unlearning}
Our node-level contrastive unlearning utilizes two key observed properties of node embeddings. %Such embeddings present  geometric patterns obtained from the training of $V_{tr}$. For a node classification model, we hypothesize that 
First, the embedding of a training node $v$ is similar to those of other nodes with the same class and distant from those of the nodes with a different class. This is supported by existing literature that empirically and mathematically showed that  the embeddings of intra-class samples are similarly located in the embedding space, and inter-class embeddings are distantly located, for a classification model trained with cross-entropy loss~\cite{das_separability_2019}. Second, the embeddings of $v$ and its neighbors are closely located in the embedding space. This phenomenon is supported by various works and investigated closely with the smoothing effect of GNNs~\cite{keriven2022not, dong2024smoothgnn}. In short, a training node $v$'s embedding is closely located with (1) embeddings of other nodes with the same class as $v$ and (2) embeddings of $v$'s neighbors.
From these observations, we achieve unlearning by modifying embeddings of unlearning node (node representation unlearning) and maintain the model utility via neighborhood reconstruction. 

\partitle{Node representation unlearning} 
Our unlearning goal is to disassociate the embeddings of the unlearning node from the embeddings of its neighbors and nodes with the same class up to the point where the model perceives the unlearning nodes as unseen nodes. To achieve this, we contrast each unlearning node with (1) randomly selected remaining nodes with different classes to pull embeddings of unlearning nodes towards them and (2) neighbors with the same class to push embeddings of unlearning nodes away from them. To this end, embeddings of unlearning nodes are steered away from embeddings of neighbors and nodes with the same class and locate near the decision boundary where test nodes' embeddings are.

In each round of unlearning, a mini-batch $B_{u} \subset V_{u}$ and its $k$-hop subgraph $G_{B_{u}} = \{B_{u}\cup \mathcal{N}^k_{B_{u}}, E^k_{B_{u}}, X^k_{B_{u}}\}$ are sampled where $\mathcal{N}^k_{B_{u}}$ is a set of $k$-hop neighbor nodes of $B_{u}$.   $X^k_{B_{u}}$ is a set of node features of $B_{u} \cup \mathcal{N}_{B_{u}}$ and $E^k_{B_{u}}$ is a set of edges of $k$-hop subgraph. Let $H_{u} = f_{\mathcal{E}}\left(B_{u}, G_{B_{u}}\right)$ be the node representations of $B_{u}$. Correspondingly, a mini-batch of remaining nodes $B_{r} \subset V_r$ and its $k$-hop subgraph $G_{B_r}$ is sampled. We denote node representations of remaining nodes by $H_r = f_{\mathcal{E}}\left(B_r, G_{B_r}\right)$. Finally, we denote $H_{nb}$ as a set of node representations of one-hop neighbors of $B_u$. For the $i$-th unlearning node $v_{i} \in V_{u}$, we compose positive  and negative representations for contrastive unlearning from $H_{r}$ and $H_{nb}$, respectively. The positive set is $P(v_{i}) = \{h_{nb, j} \vert h_{nb, j} \in H_{nb}, y_j = y_i \}$, representations of immediate neighbors of $v_{i}$ with the same class. The negative set is $N(v_{i}) = \{h_{r, j} \vert h_{r, j} \in H_r,  y_j \neq y_i\}$, representations of remaining nodes with different classes from $v_{i}$. Contrastive unlearning loss aims to minimize similarity of embeddings from positive set and maximize similarity of embeddings with negative embeddings. Specifically, the node unlearning loss $\mathcal{L}$ can be expressed as follows   
\begin{align}
    & \mathcal{L}_{U} = \sum_{v_{i} \in V_u}{\frac{-1}{\lvert N\left(v_{i}\right) \rvert}} \sum_{h_{n} \in N}{\log{\frac{\exp{\left(h_{i} \cdot h_n\right)}/\tau}{\sum\limits_{h_p \in P\left(v_{i}\right)}{\exp{\left(h_{i}\cdot h_p \right)}/\tau}}}}
    \label{eq:ul_loss}
\end{align}
where $\tau \in \mathcal{R}^+$ is a scalar temperature parameter and $h_i$ is the node embedding of $v_i$. To this end, $h_i$ is pushed towards $h_n$ and pulled away from $h_p$, effectively isolating it from embeddings of neighbors and embeddings of nodes with the same class.

\partitle{Neighborhood Reconstruction}
A fundamental difference between a GNN and a feed-forward network is the neighborhood aggregation. Embeddings of every sample are independently obtained from the feed-forward network. In contrast, embeddings of each node from a $k$-layer GNN are the aggregates of all embeddings of $k$-hop neighbors of the node. Every node of $\mathcal{N}^k_{V_u}$ is affected by embeddings of nodes of $V_u$. This means that modified embeddings of $V_u$ from the node-level contrastive unlearning stage are propagated to their neighbors during inference of the neighbors, which can reduce the model utility. Thus, to properly ensure the model utility, it is important to completely remove the influence of $V_u$ from $\mathcal{N}^k_{V_u}$ by reversing the neighborhood aggregation.

We recall the observation that the embeddings of a node are closely located with embeddings of its neighbors. 
Consider two nodes $v_i$ and $v_j$ who are neighbors and $v_k$ who is not a neighbor to either of $v_i$ or $v_j$. Due to the neighborhood aggregation, a model will generate similar embeddings for $v_i$ and $v_j$, while embeddings of $v_k$ will be dissimilar. % to embeddings of $v_i$ or $v_j$, since $v_k$ is not a neighbor of either of them. 
Accordingly, to remove the propagation of embeddings of $V_u$, a completely unlearned model should ensure the embeddings of $\mathcal{N}^k_{V_u}$ are dissimilar to the embeddings of $V_u$.

Note that the node representation unlearning as shown in equation \ref{eq:ul_loss}
%To achieve this, we propose the neighborhood reconstruction, which 
does this to some extent by pushing the unlearning nodes further away from embeddings of  their neighbors.  However, this is not sufficient for two reasons: (1) directions to push the neighbors' embeddings are unstable because embeddings of unlearning nodes are constantly changing through the unlearning process and, (2) using only the direction opposite from the unlearning nodes could cause bias. Thus, relying only on the unlearning nodes as an anchor to push away embeddings of neighbors could incorrectly steer the representation of embeddings, which can lead to ineffective disconnection and can cause utility loss. Moreover, neighbors with different class to the unlearning nodes never participate in node representation unlearning. Thus it is crucial to modify their embeddings to maintain overall model utility. 

For neighborhood reconstruction, we aim to correct embeddings of neighbors by pulling embeddings of each neighbor towards other remaining neighbors. Specifically, for all nodes in $k-1$ hop neighbors of $V_u$, we modify their embeddings by pushing them to their neighbors ($k$-hop neighbors of $V_u$) excluding $V_u$. Let $v_i \in \mathcal{N}^{k-1}_{B_u}$, we compose a negative set $S\left(v_i\right) \subseteq \mathcal{N}^k_{B_u}\backslash V_u$ where each $v_j \in S$ is a neighbor of $v_i$ and $S_H\left(v_i\right)$ as representations of nodes in $S(v_i)$. The neighborhood reconstruction maximizes similarity of $v_i$'s embedding to the embeddings of the negative set. Accordingly, the neighborhood reconstruction loss is defined as follows
\begin{align}
& \mathcal{L}_N = \sum_{v_i \in \mathcal{N}^{k-1}_{B_u}}{\frac{-1}{\lvert S\left(v_i\right)\rvert}}\sum_{h_j \in R_S\left(v_i\right)}{\frac{h_i \cdot h_j}{\tau}}
\end{align}
The loss effectively pushes embeddings of every $k-1$-hop neighbors to its remaining neighbors ($k$-hop neighbors of $V_u$). As we can see, the embeddings of closer neighbors of $V_u$ should be optimized in relation to the embeddings of further neighbors. We do not include the embeddings of $k$-th hop neighbors for the reconstruction. Instead, we only update them with cross entropy loss to stabilize their embeddings as they serve as anchors to push $k-1$ hop neighbors' embeddings. %Because optimizing embeddings of neighbors are determined by the further neighbors, 
 Also, neighborhood reconstruction recursively modifies the embeddings of neighbors, as it is crucial to modify the farthest neighbors first to ensure correct positioning of embeddings of closer neighbors. %Thus we recursively modify the embeddings to effectively and properly optimize the neighbors.

%Similar to previous sections, we utilize cross-entropy loss in conjunction with reconstruction loss for further stabilization of representations space.

\partitle{Cross entropy loss}
% The authors of contrastive unlearning~\cite{zhang2024contrastive}, highlighted that contrasting the embeddings of unlearning samples inadvertently affects the embeddings of remaining samples. 
To further stabilize the model utility, 
%Repeating the above unlearning process inevitably changes the embeddings of remaining nodes and can potentially reduce model performance. %The original framework effectively mitigated this by introducing a cross-entropy loss for remaining samples to keep the representations of remaining samples to their original position. 
we use a similar idea as ~\cite{zhang2024contrastive}  and add an auxiliary cross-entropy loss for both node representation unlearning and neighborhood reconstruction. and update all remaining nodes involved in both methods. The total contrastive unlearning loss is as follows.
% The authors of contrastive unlearning~\cite{zhang2024contrastive}, highlighted that contrasting the embeddings of unlearning samples inadvertently affects the embeddings of remaining samples. Repeating this process inevitably reduces model performance. The original framework effectively mitigated this by introducing a cross-entropy loss for remaining samples to keep the representations of remaining samples in their original position. We follow this idea and use the cross-entropy loss for remaining nodes to stablize their embeddings. The total unlearning loss is as follows.
 \begin{align}
     & \mathcal{L}_{\textrm{Node}} = \mathcal{L}_U + \beta\mathcal{L}_C\left(f\left(B_{rem}\right), Y_{B_{rem}}\right)
 \end{align}
Where $\beta$ is a hyperparameter to determine weights for each loss term, $\mathcal{L}_C$ is the cross-entropy loss, $B_{rem}$ is a batch sampled from $V_{rem}$, and $Y_{B_{rem}}$ is the label set of $B_{rem}$.

For neighborhood reconstruction, we apply cross-entropy loss for all neighbors. The total neighborhood reconstruction loss is as follows.
\begin{align}
    & \mathcal{L}_{\textrm{Neighbor}} = \mathcal{L}_N + \gamma\mathcal{L}_C\left(f\left(v_i\right), y_i\right)
\end{align}
Where $\gamma$ is a hyperparameter to determine weights for each loss term, $v_i\in N^{k-1}_{V_u}$ is a node of $k-1$-hop neighbors of $V_u$.

\partitle{Termination Condition}
A remaining challenge is to determine the right moment to terminate the unlearning process. Stopping too early would cause insufficient unlearning, and stopping too late would overly modify the embedding space, causing a detrimental effect on model performance. We design an explicit termination condition to achieve good model performance and effective unlearning. We assume a subset of nodes $V_{eval} \subseteq V_{ts}$ and a subgraph consisting of $V_{eval}$ are available for determining the termination condition.
Recall our problem definition of~\ref{eq:prob1}. If a model achieves higher accuracy for $V_u$ than accuracy on unseen test nodes, it indicates that it possesses inherent knowledge about $V_u$. Therefore, to ensure that the model does not retain knowledge of $V_u$, we aim to reduce the accuracy for $V_u$ to be no greater than that for $V_{eval}$, which are essentially unseen nodes. Accordingly, we design the algorithm to terminate as soon as it satisfies the following termination condition:
\begin{align}
    & \mathbf{Acc}\left(f^\prime\left(V_{u}\right), Y_{u}\right) \leq \mathbf{Acc}\left(f^\prime\left(V_{eval}\right), Y_{eval}\right)
    \label{eq:termination}
\end{align}
Terminating the algorithm before satisfying condition~\ref{eq:termination} would leave inherent knowledge of $V_u$ within the model, resulting in insufficient unlearning. In addition, it is not desired to continue after achieving the condition because it forcefully steers $f^\prime$ to deliberately make false predictions on $V_u$, which is not aligned with our goal of unlearning and can be exploited  to infer the membership of $V_u$.

\partitle{Full algorithm}
The entire algorithm sequentially processes node representation unlearning and neighborhood reconstruction. Refer to Appendix~\ref{sec:full_algo} for the detailed illustration on the full-algorithm.

\section{Experiments}
\subsection{Setup}
\partitle{Dataset and Models}
We use four benchmark datasets: Cora-ML, PubMed, Citeseer and CS, and employ Graph Convolutional Networks (GCN)~\cite{kipf2016semi}, Graph Attention Network (GAT)~\cite{velivckovic2017graph}, and Graph Isomorphism network (GIN)~\cite{xu2018powerful} for comparison. Performance of each model of each dataset is in Appendix~\ref{sec:implement_hyperparam}. 
%Readers may refer to Appendix~\ref{sec:code} for implementations (code).
We provide our code at \textcolor{blue}{\href{https://anonymous.4open.science/r/Node-CUL-E30D/}{an anonymized git repository}}.

We randomly select 10\% of nodes from a graph data as test nodes and 90\% of the nodes as training nodes. Also, we select 10\% of training nodes as unlearning nodes. As we use a transductive setting, test nodes can be accessed by the GNN during the forward pass; however, they are not used during the optimization. 

\begin{table*}[!h]
    \centering
    \small
    \setlength{\tabcolsep}{1pt}
    \begin{tabularx}{\textwidth}{ll>{\centering\arraybackslash}X>{\centering\arraybackslash}X>{\centering\arraybackslash}X>{\centering\arraybackslash}X>{\centering\arraybackslash}X>{\centering\arraybackslash}X>{\centering\arraybackslash}m{0.05\textwidth}>{\centering\arraybackslash}m{0.05\textwidth}>{\centering\arraybackslash}m{0.05\textwidth}}
        \toprule
        & & \multicolumn{3}{c}{\scalebox{0.9}{Test accuracy $\uparrow$}} & \multicolumn{3}{c}{\scalebox{0.9}{Unlearn accuracy}} & \multicolumn{3}{c}{\scalebox{0.9}{Unlearn score $\downarrow$}} \\
        \cmidrule(lr){3-5} \cmidrule(lr){6-8} \cmidrule(lr){9-11}
        \scalebox{0.9}{Dataset} & \scalebox{0.9}{Method} & \scalebox{0.9}{GCN} & \scalebox{0.9}{GAT} & \scalebox{0.9}{GIN} 
        & \scalebox{0.9}{GCN} & \scalebox{0.9}{GAT} & \scalebox{0.9}{GIN}
        & \scalebox{0.9}{GCN} & \scalebox{0.9}{GAT} & \scalebox{0.9}{GIN} \\
        \midrule
        \multirow{4}{*}{\scalebox{0.9}{Cora-ML}}
        & \scalebox{0.9}{Retrain (Reference)} & 87.16$\pm$0.17 & 86.07$\pm$0.64 & 85.05$\pm$1.22 & 84.31$\pm$0.31 & 85.00$\pm$1.20 & 84.72$\pm$1.36 & 2.84 & 1.07 & 0.34\\ 
        & \scalebox{0.9}{Node-CUL} & \textbf{87.65$\pm$1.42} & \textbf{86.66$\pm$0.79} & \textbf{87.90$\pm$1.14} & 85.03$\pm$0.577 & 81.94$\pm$0.75 & 85.80$\pm$1.43 & \textbf{2.62} & \textbf{4.72} & \textbf{2.1}\\
        & \scalebox{0.9}{GNNDelete} & 85.92$\pm$0.30 & 85.22$\pm$0.86 & 86.17$\pm$0.69 & 28.77$\pm$5.21 & 6.63$\pm$1.52 & 30.71$\pm$3.16 & 51.75 & 78.59 & 55.46\\
        & \scalebox{0.9}{GIF} & 84.69$\pm$1.06 & 80.86$\pm$0.17 & 62.87$\pm$35.21 & 92.18$\pm$1.67 & 90.74$\pm$1.00 & 66.82$\pm$38.41 & 7.49 & 9.88 & 3.95\\
        \midrule
        \multirow{4}{*}{\scalebox{0.9}{PubMed}}
        & \scalebox{0.9}{Retrain (Reference)} & 88.872$\pm$0.06 & 88.494$\pm$0.18 & 88.753$\pm$0.35 & 87.40$\pm$0.73 & 86.66$\pm$0.18 & 88.01$\pm$0.14 & 1.47 & 1.83 & 0.74 \\
        & \scalebox{0.9}{Node-CUL} & \textbf{89.09$\pm$0.07} & \textbf{87.97$\pm$0.21} & \textbf{88.22$\pm$0.45} & 86.83$\pm$0.41 & 86.36$\pm$0.22 & 85.94$\pm$0.20 & 2.26 & \textbf{1.61} & 2.28\\
        & \scalebox{0.9}{GNNDelete} & 85.86$\pm$0.26 & 85.67$\pm$0.16 & 86.47$\pm$0.15 & 39.69$\pm$0.93 & 38.49$\pm$0.32 & 39.69$\pm$0.93 & 46.26 & 47.18 & 46.78\\
        & \scalebox{0.9}{GIF} & 84.93$\pm$0.08 & 86.67$\pm$0.06 & 72.73$\pm$20.22 & 86.15$\pm$0.65 & 88.54$\pm$0.10 & 72.45$\pm$22.18 & \textbf{1.22} & 1.87 & \textbf{0.28}\\
        \midrule
        \multirow{4}{*}{\scalebox{0.9}{Citeseer}}
        & \scalebox{0.9}{Retrain (Reference)} & 77.91 $\pm$ 1.11 & 77.91$\pm$0.99 & 78.11 $\pm$0.61 & 72.43$\pm$0.71 & 74.06$\pm$0.61 & 73.43$\pm$2.16 & 5.48 & 3.85 & 4.67 \\
        & \scalebox{0.9}{Node-CUL} & \textbf{78.21$\pm$0.37} & 77.51$\pm$0.75 & \textbf{78.78$\pm$0.65} & 71.92$\pm$2.55 & 73.55$\pm$0.98 & 77.86$\pm$0.74 & \textbf{6.29} & \textbf{3.96} & \textbf{0.92}\\
        & \scalebox{0.9}{GNNDelete} & 76.90$\pm$0.37 & 76.40$\pm$0.75 & 78.41$\pm$0.37 & 22.18$\pm$1.62 & 22.18$\pm$1.62 & 22.18$\pm$1.63 & 54.72 & 54.22 & 56.23\\
        & \scalebox{0.9}{GIF} & 76.90$\pm$0.51 & \textbf{78.01$\pm$0.64} & 77.61$\pm$0.99 & 87.46$\pm$1.07 & 85.83$\pm$1.24 & 85.71$\pm$1.91 & 10.56 & 7.82 & 8.1\\
        \midrule
        \multirow{3}{*}{\scalebox{0.9}{CS}} 
        & \scalebox{0.9}{Retrain (Reference)} & 91.45$\pm$0.53 & 93.40$\pm$0.97&89.31$\pm$0.19&87.62$\pm$1.06&89.88$\pm$0.61&84.58$\pm$0.48&3.82&3.52 & 4.72 \\
        & \scalebox{0.9}{Node-CUL} & \textbf{94.59$\pm$0.25} & \textbf{95.81$\pm$0.16} & \textbf{90.14$\pm$0.10} & 92.30$\pm$0.61 & 93.96$\pm$0.46 & 89.88$\pm$0.18 & 2.29 & \textbf{1.85} & \textbf{0.26}\\
        & \scalebox{0.9}{GNNDelete} & 92.31$\pm$0.61 & 93.96$\pm$0.46 & 89.88$\pm$0.18 & 16.18$\pm$0.59 & 9.66$\pm$3.29 & 72.26$\pm$0.37 & 76.13 & 84.3 & 17.62\\
        & \scalebox{0.9}{GIF} & 94.47$\pm$0.02 & 95.37$\pm$0.01 & 85.59$\pm$4.01 & 94.54$\pm$0.24 & 92.61$\pm$0.37 & 75.56$\pm$8.23 & \textbf{0.07} & 2.76 & 10.03\\
        \bottomrule
    \end{tabularx}
    \caption{Performance evaluation on different datasets.}
    \label{tab:utility}
    \vspace{-1em}
\end{table*}

\begin{table}[!h]
    \centering
    \small
    \setlength{\tabcolsep}{1pt}
    \begin{tabularx}{\linewidth}{*{5}{>{\centering\arraybackslash}X}}
        \toprule
        \scalebox{0.9}{Dataset} & & \scalebox{0.9}{GCN} & \scalebox{0.9}{GAT} & \scalebox{0.9}{GIN} \\
        \midrule
        \multirow{4}{*}{\scalebox{0.9}{CoraML}}
        & \scalebox{0.9}{Retrain (Ref.)} & 0.4899 & 0.4899 & 0.5185 \\
        & \scalebox{0.9}{\textbf{Node-CUL}} & 0.4768 & 0.4655 & 0.4844\\
        & \scalebox{0.9}{GNNDelete} & 0.3705 & 0.3703 & 0.3678 \\
        & \scalebox{0.9}{GIF} & 0.5181 & 0.5278 & 0.5455\\
        \midrule
        \multirow{4}{*}{\scalebox{0.9}{PubMed}} 
        & \scalebox{0.9}{Retrain (Ref.)} & 0.5103 & 0.4972 & 0.5053 \\
        & \scalebox{0.9}{\textbf{Node-CUL}} & 0.4883 & 0.4850 & 0.4963 \\
        & \scalebox{0.9}{GNNDelete} & 0.4317 & 0.4449 & 0.4203 \\
        & \scalebox{0.9}{GIF} & 0.5001 & 0.4867 & 0.5052 \\
        \midrule
        \multirow{4}{*}{\scalebox{0.9}{Citeseer}}
        & \scalebox{0.9}{Retrain (Ref.)} & 0.4684 & 0.4677 & 0.4582 \\
        & \scalebox{0.9}{\textbf{Node-CUL}} & 0.5019 & 0.4867 & 0.4858\\
        & \scalebox{0.9}{GNNDelete} & 0.3955 & 0.3778 & 0.3644\\
        & \scalebox{0.9}{GIF} & 0.5422 & 0.5279 & 0.5310\\
        \midrule
        \multirow{4}{*}{\scalebox{0.9}{CS}} 
        & \scalebox{0.9}{Retrain (Ref.)} & 0.4608 & 0.4050 & 0.4864 \\
        & \scalebox{0.9}{\textbf{Node-CUL}} & 0.4867 & 0.5171 & 0.4760\\
        & \scalebox{0.9}{GNNDelete} & 0.5036 & 0.6325 & 0.4791 \\
        & \scalebox{0.9}{GIF} & 0.4780 & 0.3916 & 0.4753 \\
        \bottomrule
    \end{tabularx}
    \caption{AUC of LiRA detection performance on CoraML and PubMed datasets}
    \label{tab:mia_datasets}
    \vspace{-1em}
\end{table}

\partitle{Comparison Methods} 
We include three baseline methods for GNN unlearning: 1) \textbf{Retrain} is fully re-training a GNN model with remaining nodes only, and it serves as a reference of a perfectly unlearned model to compare the unlearning effectiveness and utility. 2) \textbf{Graph Influence Function (GIF)}~\cite{wu2023gif} captures the influence of a node or an edge to unlearn spanning through its $k$-hop neighbors and conducts a one-shot update to remove the influence. We utilize their node-unlearning framework. 3) \textbf{GNNDelete}~\cite{cheng2023gnndelete} inserts unlearning layers in between GNN layers and optimizes the unlearning layers for two loss terms: delete-edge consistency and neighborhood influence. The former ensures complete deletion of target edges, and the latter reduces the impact of deletion throughout subgraphs consisting of the edges.
While there are other SISA-based GNN unlearning frameworks based on partitioning and efficient retraining, we do not compare them with ours as these works achieve unlearning in fundamentally different ways; hence, it is difficult to directly compare the results. We aim to compare ours and SOTA frameworks that rely on optimizing model parameters for unlearning.

For every experiment, we provide the average and standard deviation of three runs with different seeds. Also, we conduct experiments with the best hyperparameters for Node-CUL and baselines. Refer to Appendix~\ref{sec:implement_hyperparam} for detailed hyperparameter settings.

\partitle{Evaluation Metrics}
1) \textbf{Model performance.} We evaluate the test accuracy of $V_{ts}$ from unlearned models. 2) \textbf{Unlearn efficacy.} We assess accuracy on $V_u$ (unlearning nodes) and compare it with the accuracy of $V_{ts}$ (test nodes). A successfully unlearned model should exhibit similar accuracy for both unlearning and test nodes. We provide a metric of unlearn score, which is the absolute difference between the accuracy of test and unlearning nodes~\cite{zhang2024contrastive}. 3) \textbf{Efficiency.} We measure the runtime of each unlearning framework.

\partitle{Verifying Unlearning via Membership Inference Attack} We conduct a membership inference attack on unlearned models to evaluate the effectiveness of unlearning from different frameworks. We re-purpose the likelihood ratio attack (LiRA)~\cite{carlini2022membership}. We mark the entire unlearning nodes as members and randomly select the same number of test nodes as non-members. Then we train 32 shadow models using the original datasets and test the likelihood that the unlearning nodes were part of the training nodes.  We report AUC values and AUROC curves. A successfully unlearned model should have difficulty discerning unlearning nodes as members, hence an AUC close to 0.5 which is equivalent to a random guess.

\begin{figure*}[ht]
\small
    \centering
    % First row - GCN
    \begin{subfigure}[t]{0.23\textwidth}
        \centering
        \includegraphics[width=\textwidth]{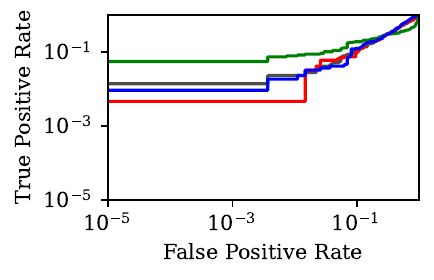}
        \vspace{-2em}  % Reduce space between image and caption
        \caption{GCN - CoraML}
        \label{fig:auroc_gcn_cora}
    \end{subfigure}
    \hfill
    \begin{subfigure}[t]{0.23\textwidth}
        \centering
        \includegraphics[width=\textwidth]{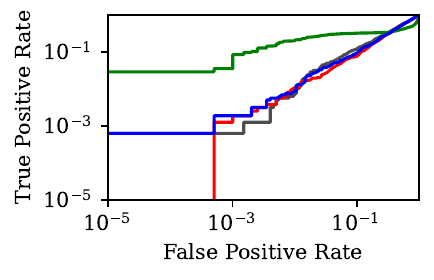}
        \vspace{-2em}
        \caption{GCN - PubMed}
        \label{fig:auroc_gcn_pubmed}
    \end{subfigure}
    \hfill
    \begin{subfigure}[t]{0.23\textwidth}
        \centering
        \includegraphics[width=\textwidth]{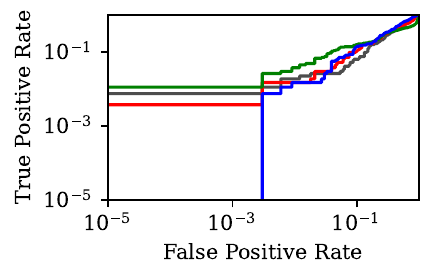}
        \vspace{-2em}
        \caption{GCN - Citeseer}
        \label{fig:auroc_gcn_citeseer}
    \end{subfigure}
    \begin{subfigure}[t]{0.23\textwidth}
        \centering
        \includegraphics[width=\textwidth]{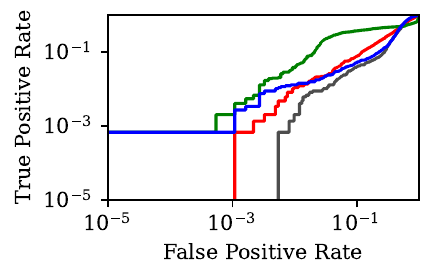}
        \vspace{-2em}
        \caption{GCN - CS}
        \label{fig:auroc_gcn_cs}
    \end{subfigure}
    
    \vspace{1em}  % Add space between rows
    
    % Second row - GAT
    \begin{subfigure}[t]{0.23\textwidth}
        \centering
        \includegraphics[width=\textwidth]{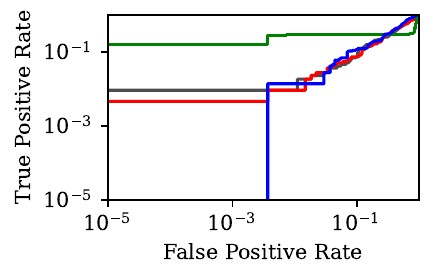}
        \vspace{-2em}
        \caption{GAT - CoraML}
        \label{fig:auroc_gat_cora}
    \end{subfigure}
    \hfill
    \begin{subfigure}[t]{0.23\textwidth}
        \centering
        \includegraphics[width=\textwidth]{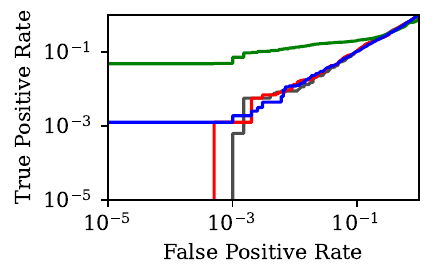}
        \vspace{-2em}
        \caption{GAT - PubMed}
        \label{fig:auroc_gat_pubmed}
    \end{subfigure}
    \hfill
    \begin{subfigure}[t]{0.23\textwidth}
        \centering
        \includegraphics[width=\textwidth]{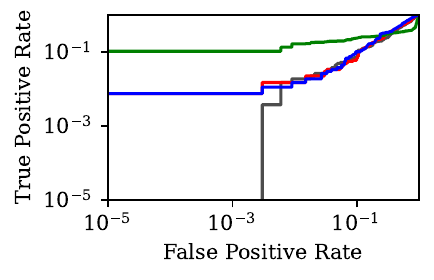}
        \vspace{-2em}
        \caption{GAT - Citeseer}
        \label{fig:auroc_gat_citeseer}
    \end{subfigure}
    \begin{subfigure}[t]{0.23\textwidth}
        \centering
        \includegraphics[width=\textwidth]{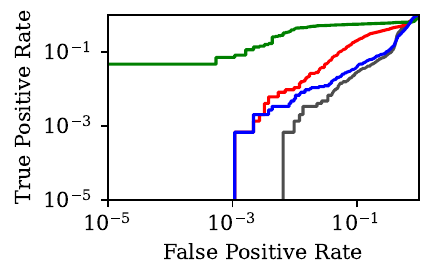}
        \vspace{-2em}
        \caption{GAT - CS}
        \label{fig:auroc_gat_cs}
    \end{subfigure}
    
    \vspace{1em}  % Add space between rows
    
    % Third row - GIN
    \begin{subfigure}[t]{0.23\textwidth}
        \centering
        \includegraphics[width=\textwidth]{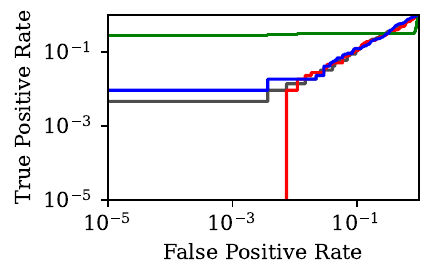}
        \vspace{-2em}
        \caption{GIN - CoraML}
        \label{fig:auroc_gin_cora}
    \end{subfigure}
    \hfill
    \begin{subfigure}[t]{0.23\textwidth}
        \centering
        \includegraphics[width=\textwidth]{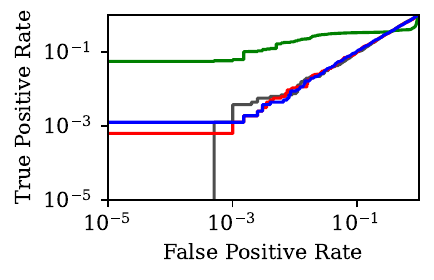}
        \vspace{-2em}
        \caption{GIN - PubMed}
        \label{fig:auroc_gin_pubmed}
    \end{subfigure}
    \hfill
    \begin{subfigure}[t]{0.23\textwidth}
        \centering
        \includegraphics[width=\textwidth]{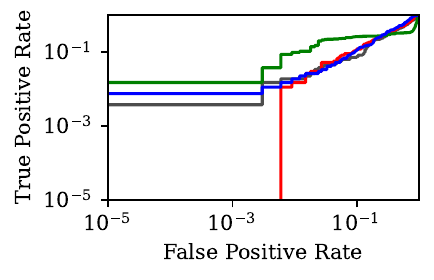}
        \vspace{-2em}
        \caption{GIN - Citeseer}
        \label{fig:auroc_gin_citeseer}
    \end{subfigure}
    \begin{subfigure}[t]{0.23\textwidth}
        \centering
        \includegraphics[width=\textwidth]{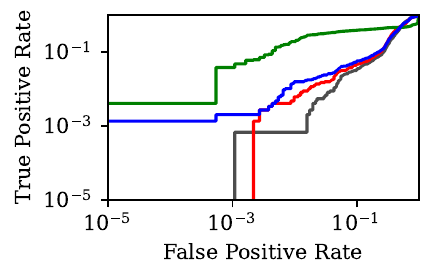}
        \vspace{-2em}
        \caption{GIN - CS}
        \label{fig:auroc_gin_cs}
    \end{subfigure}
    \begin{subfigure}[t]{0.5\textwidth} 
        \centering
        \includegraphics[width=\textwidth]{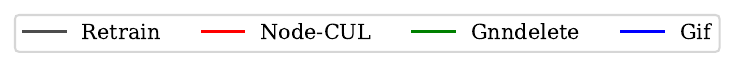}
        \vspace{-1em}
        \label{fig:bottom_figure}
    \end{subfigure}
    \vspace{-1em}
    \caption{AUROC curves of LiRA's detection performance on different models and datasets}
    \label{fig:auroc_all}
\end{figure*}

\begin{table}[ht]
    \centering
    \small
    \setlength{\tabcolsep}{0.5pt}
    \begin{tabularx}{\linewidth}{l>{\centering\arraybackslash}X*{3}{>{\centering\arraybackslash}X}}
        \toprule
        Dataset & Method & GCN & GAT & GIN \\
        \midrule
        \multirow{4}{*}{CoraML} 
        & Retrain (Ref.) & 64.82$\pm$10.20 & 58.21$\pm$0.44 & 51.83$\pm$4.44 \\
        & \textbf{Node-CUL} & 29.39$\pm$8.40 & 30.03$\pm$4.37 & 59.19$\pm$9.99 \\
        & GNNDelete & 8.59$\pm$2.20 & 28.43$\pm$2.011 & 7.23$\pm$1.38 \\
        & GIF & 24.04$\pm$8.46 & 50.42$\pm$12.38 & 47.27$\pm$11.71 \\
        \midrule
        \multirow{4}{*}{PubMed}
        & Retrain (Ref.) & 391.65$\pm$67.45 & 348.98$\pm$11.64 & 273.42$\pm$11.66 \\
        & \textbf{Node-CUL} & 101.35$\pm$12.25 & 165.77$\pm$3.79 & 149.38$\pm$50.65 \\
        & GNNDelete & 79.50$\pm$1.27 & 194.71$\pm$0.51 & 79.05$\pm$4.027 \\
        & GIF & 41.95$\pm$8.42 & 30.07$\pm$0.03 & 30.35$\pm$0.46 \\
        \midrule
        \multirow{4}{*}{Citeseer}
        & Retrain (Ref.) & 100.36$\pm$4.73 & 92.41$\pm$8.39& 78.38$\pm$7.40\\
        & \textbf{Node-CUL} & 21.34$\pm$7.04 & 43.04$\pm$7.32 & 81.14$\pm$6.14 \\
        & GNNDelete & 24.08$\pm$0.62 & 62.17$\pm$4.22 & 31.18$\pm$0.57 \\
        & GIF & 28.37$\pm$2.25 & 30.05$\pm$0.02 & 37.03$\pm$5.13 \\
        \midrule
        \multirow{4}{*}{CS}
        & Retrain (Ref.) & 6232.1$\pm$454.1& 6139.1$\pm$705.8&733.8$\pm$884.6\\
        & \textbf{Node-CUL} & 256.18$\pm$32.61 & 257.05$\pm$7.86 & 166.72$\pm$8.93 \\
        & GNNDelete & 250.76$\pm$14.09 & 277.20$\pm$27.14 & 203.49$\pm$14.44 \\
        & GIF & 77.11$\pm$13.23 & 100.17$\pm$25.17 & 40.13$\pm$14.36 \\
        \bottomrule
    \end{tabularx}
    \caption{Running time of unlearning framework on different datasets (seconds)}
    \label{tab:time_datasets}
    \vspace{-1em}
\end{table}

\begin{table*}[h]
    \centering
    \small
    \begin{tabular}{c|ccc|ccc|ccc}
        \toprule
        & \multicolumn{3}{c|}{Test Accuracy $\uparrow$} & \multicolumn{3}{c}{Unlearn Accuracy} & \multicolumn{3}{c}{Unlearn Score $\downarrow$} \\
        \cmidrule(lr){2-4} \cmidrule(lr){5-7} \cmidrule(lr){8-10}
        \textbf{Ratio} & \textbf{Node-CUL} & \textbf{GNNDelete} & \textbf{GIF} & \textbf{Node-CUL} & \textbf{GNNDelete} & \textbf{GIF} & \textbf{Node-CUL} & \textbf{GNNDelete} & \textbf{GIF} \\
        \midrule
        10\% & \textbf{87.65$\pm$1.42}& 85.92$\pm$0.30& 84.69$\pm$1.06& 85.03$\pm$0.57 &28.77$\pm$5.21 & 92.18$\pm$1.67&\textbf{2.62}& 51.75& 7.49\\ 
        20\% & \textbf{87.53$\pm$0.46}& 85.19$\pm$0.60& 84.64$\pm$0.67& 86.44$\pm$0.46 &31.41$\pm$1.99 & 93.84$\pm$1.63&\textbf{1.09}& 53.77& 9.19\\
        30\% & \textbf{87.16$\pm$0.63}& 84.32$\pm$0.76& 84.32$\pm$1.77& 84.77$\pm$0.45 &30.77$\pm$1.27 & 91.44$\pm$2.67&\textbf{2.39}& 53.55& 7.12\\
        40\% & \textbf{86.73$\pm$0.61}& 82.96$\pm$0.80& 84.48$\pm$1.16& 84.95$\pm$0.66 &31.49$\pm$1.37 & 93.27$\pm$1.19&\textbf{1.78}& 51.47& 8.79\\
        50\% & \textbf{86.29$\pm$0.60}& 80.98$\pm$0.46& 83.09$\pm$1.15& 84.94$\pm$0.87 &32.17$\pm$0.72 & 91.75$\pm$2.22&\textbf{1.36}& 48.82& 8.67\\
        60\% & \textbf{86.29$\pm$1.32}& 80.62$\pm$0.87& 83.58$\pm$0.92& 84.89$\pm$1.13 &31.75$\pm$0.59 & 94.41$\pm$0.14&\textbf{1.39}& 48.87& 10.83\\
        \bottomrule
    \end{tabular}
    \caption{Performance comparison with different unlearning ratios}
    \label{tab:ratio_comparison}
    \vspace{-1em}
\end{table*}

\subsection{Model Utility}
\label{sec:exp_model_utility}
Table~\ref{tab:utility} shows test accuracy, unlearn accuracy, and unlearn score of different methods on various datasets and GNN models. A successful unlearning framework should minimize the utility loss of the resulting unlearned model. From the table, Node-CUL achieves the best test accuracy for most of the datasets and models. For the Cora-ML dataset, the test accuracy of the unlearned model from Node-CUL is even higher than the test accuracy of the original model. This is likely attributed to neighborhood reconstruction. As it optimizes neighbors of unlearning nodes, it gradually enhances prediction performance. The performance of the unlearned model from GNNDelete is almost similar to the original model. GNNDelete activates unlearning layers only for unlearning nodes and deactivates them for all other nodes. Thus, most of the test nodes are processed through the original GNN layers, preserving the test accuracy. However, it is highly impractical and questionable to still keep track of unlearning nodes for inference after the unlearning process. % While GIF shows stable test accuracy across all datasets, 
The test accuracy of GIF is consistently lower than Node-CUL. More importantly, GIF often fails to preserve model utility for GIN models. Overall, Node-CUL shows the highest performance across all models and datasets.

The unlearn score indicates the unlearn efficacy. As we hypothesized from section~\ref{sec:prob_def}, perfectly unlearned models (retrain) show a small difference in test and unlearn accuracy, resulting in small unlearn scores. Accordingly, a successfully unlearned model should show similar test and unlearn accuracy, or a low unlearn score. Node-CUL demonstrated a low unlearn score across all models and datasets. In contrast, GNNDelete shows a very high unlearn score, due to very low unlearn accuracy. Unlearning layers of GNNDelete are optimized to make a randomized prediction for unlearning nodes. Thus, when enabled, the layers destroy embeddings of unlearning nodes. However, as we have mentioned earlier, having a very low unlearn accuracy or high unlearn score could be problematic as it indicates that the model behaves differently on unlearning and test samples. This difference can be exploited by  membership inference attacks, further increasing the privacy risk. GIF shows a relatively smaller unlearn score, indicating that it is somewhat effective in unlearning. However, its unlearn accuracy tends to be higher than test accuracy for most of the datasets. It implies that the model still retains some knowledge of unlearning nodes. Overall, Node-CUL is showing the lowest unlearn score for most of the models and datasets with its unlearn accuracy consistently lower than the test accuracy.

\subsection{Unlearn Efficacy via MIA}
Table~\ref{tab:mia_datasets} shows the AUC of LiRA attack on the unlearned models with different unlearning frameworks. We omit standard deviations as they are negligible. Similar to retrained models, a successfully unlearned model should present the attack AUC close to 0.5. Notably, GNNDelete is showing a very low AUC, usually around 0.38. This occurs because the attack misclassified unlearning nodes as non-member nodes and vice versa. This is problematic because it shows that the attack is able to distinguish the unlearning and test samples. Effectively, the risk of privacy is equivalent to the case where AUC is around 0.62. As we have mentioned in the previous paragraph and Table~\ref{tab:utility}, GNNDelete had a larger unlearn score and resulted in increased privacy risk.

In contrast, the AUC of GIF and Node-CUL is close to 0.5, indicating that both methods have effectively removed the influence of unlearning nodes. The AUC of the attack on Node-CUL is mostly slightly below 0.5, which aligns closely with the AUC of the retrained model. Intuitively, it implies that the attack model was mostly making a random guess over unlearning nodes and often predicted them as non-members. This can be attributed to the termination condition of Node-CUL. The algorithm terminates as soon as unlearn accuracy drops below the test accuracy. This results in the GNN making slightly less confident predictions for unlearning nodes than test nodes. The attack mistakenly identified some unlearning nodes with low-confidence logits as non-member nodes, and some test nodes with high-confidence logits as member nodes.

While both GIF and Node-CUL show AUC close to 0.5, the key difference lies in the low false positive regime. It has been emphasized that AUC alone is not an effective metric because it does not show how confident the attack is~\cite{carlini2022membership}. From an ROC curve, having a high true positive rate when the false positive rate is higher than 50\% is not useful, as it means that the attack model is mostly predicting samples as members with low confidence. Instead, it is important to inspect the low false positive rate regime because that is where the attack model is very confident in discerning member and non-member samples. %Having a low true positive rate under the low false positive rate regime means that although the model's decision is confident, it is unsuccessful in correctly discerning the member nodes.
In the unlearning perspective, effective unlearning should prevent the attack model from successfully identifying unlearning nodes as members even when the attack model is very confident. Thus, successful unlearning should achieve a lower true positive rate when the false positive rate is very small. 

We compare ROC curves of the LiRA attack, especially in the low false positive regime in Figure~\ref{fig:auroc_all}. For the most part, Node-CUL achieves the lowest true positive rate when the false positive rate is very small. GNNDelete is showing a high true positive rate, indicating that the attack model was able to identify some unlearning nodes with high certainty. While GIF also has similarly low true positive rates for unlearning samples, it was outperformed by Node-CUL for most of the cases. Especially, Node-CUL showed a very small true positive rate for the GIN model. It clearly shows that Node-CUL achieves better unlearn efficacy and effectively removes the influence of unlearning nodes.

\subsection{Efficiency}
Table~\ref{tab:time_datasets} shows the running time (seconds) of each unlearning framework on each dataset. Note that Node-CUL potentially requires more computations than GNNDelete and GIF. GNNDelete freezes the original GNN and only optimizes unlearning layers, and GIF conducts a one-shot update for the entire GNN, while Node-CUL requires multiple updates on the entire parameters. Despite this difference, Node-CUL shows similar efficiency with GIF for the Cora-ML dataset. Node-CUL requires more computation for denser graphs. However, when unlearning the most dense graph (CS), Node-CUL was able to achieve better efficiency than GNNDelete. This is due to the termination condition, as Node-CUL was able to achieve the condition just after one unlearning round. Overall, Node-CUL incurs comparable or slightly higher computation cost than SOTA methods as a tradeoff for significantly more effective unlearning and better model utility.

\subsection{Unlearning a large number of nodes}
Table~\ref{tab:ratio_comparison} shows performance evaluation on unlearning a larger number of samples. We conduct unlearning on 10\% to 60\% of the original training data of the Cora-ML dataset and assessed test accuracy, unlearn accuracy, and unlearn score. Node-CUL achieves the best model performance across multiple ratios of unlearning. Also, it achieves the lowest unlearn score for all settings. Node-CUL is more robust in unlearning a larger number of samples, since it can leverage neighborhood reconstruction. In contrast, GNNDelete suffers utility loss as the ratio increases because more edges are involved in unlearning. When the number of unlearning nodes is small, only a small number of test nodes that have edges with unlearning nodes are processed through the unlearning layers. When the number of unlearning nodes is large, more test nodes are processed through the unlearning layers, decreasing the performance. Finally, GIF shows stable test accuracy; however, it also shows relatively high unlearn accuracy, implying ineffective unlearning. 

\subsection{Effects of neighborhood reconstruction}
\begin{table}[t]
    \centering
    \small
    \begin{tabularx}{\linewidth}{l|*{3}{>{\centering\arraybackslash}X}}
        \toprule
        & Test acc.$\uparrow$ & Unlearn acc. & Unlearn score $\downarrow$ \\
        \midrule
        With & \textbf{87.16$\pm$0.63} & 84.77$\pm$0.45 & 2.39 \\
        Without & 83.08$\pm$1.43 & 82.09$\pm$1.10 & \textbf{0.99} \\
        \bottomrule
    \end{tabularx}
    \caption{Performance evaluation of unlearning 30\% of Cora-ML dataset from the GCN model with and without the neighborhood reconstruction.}
    \label{tab:neighborhood_03}
\end{table}

We conduct an ablation study on neighborhood reconstruction to assess its model utility gain. Table~\ref{tab:neighborhood_03} shows the results of unlearning 30\% of training nodes of Cora-ML dataset from a GCN with and without the neighborhood reconstruction. The results show that having neighborhood reconstruction significantly increased the model utility with negligible loss in unlearn efficacy.  This clearly shows that neighborhood aggregation is effective for maintaining model utility.  Refer Appendix~\ref{sec:appdix_exp} for additional ablation studies.

\section{Conclusion}

In this paper, we proposed a novel node-level graph contrastive unlearning framework. It achieves unlearning by directly utilizing node embeddings from the representation space. Specifically, it utilizes contrastive loss for both node representation unlearning which adjusts the embeddings of unlearning nodes towards unseen nodes and neighborhood reconstruction which modifies the embedding
of all neighbors of the unlearning nodes to ensure the complete removal of the influences. %Then we utilize our neighborhood reconstruction, an intuitive approach to effectively reconfigure embeddings of neighbors. 
Through extensive experiments, we demonstrated that  Node-CUL  is superior to the state-of-the-art graph unlearning frameworks. In the future, we aim to extend the work for edge unlearning and general unlearning of both nodes and edges.

\bibliographystyle{ACM-Reference-Format}
\bibliography{sample-sigconf}

\pagebreak
\appendix

\section*{Appendix}
In this appendix, Section~\ref{sec:full_algo} illustrates the full algorithm of our framework. Section~\ref{sec:implement_hyperparam}  discusses detailed hyperparameter settings for experiments. Section~\ref{sec:appdix_exp} presents additional experiments.

\section{Full Algorithm}
\label{sec:full_algo}
\begin{algorithm}[]
\caption{Node-CUL}
\small
\begin{algorithmic}[1]
    \Require $f$, $f_{\mathcal{E}}(\cdot)$, $G$ \\
    \textbf{Output} $f^\prime$
    \State $U = \{(B_u, G_{B_u}) \vert G_{B_u}\subset G, \textrm{ } \forall B_u \subset V_u\}$
    \State $R = \{(B_r, G_{B_r}) \vert G_{B_r}\subset G, \textrm{ } \forall B_r \subset V_r\}$
    \While{Termination condition is not satisfied}
    \For{each $(B_u, G_{B_u}) \in U$}
            \For{$1, \cdots, \omega$}
                \State Sample $(B_r, G_{B_r}) \in R$
                \State $N = \{\mathcal{N}^1_{B_u}, \cdots , \mathcal{N}^{k+1}_{B_u}\}$, $G_N = \{G_{n}  \vert \textrm{ } \forall n\in N \}$
                \State $f \leftarrow \textrm{\textsc{Node\_Representation\_Unlearn}}\left(f, B_u, G_u, B_r, G_r\right)$
            \EndFor
            \For{$1, \cdots, \omega/2$}
            \State $f \leftarrow \textrm{\textsc{Neighborhood\_Reconstruction}}\left(f, N, G_N\right)$
            \EndFor
    \EndFor
    \State Evaluate, get termination condition with $V_{\mathrm{eval}}$
    \EndWhile \\
\textbf{return} $f^\prime$
\end{algorithmic}
\label{alg:node-cul}
\end{algorithm}

The algorithm~\ref{alg:node-cul} shows the enitre Node-CUL algorithm. $U$ is a set of $(B_u,G_{B_u})$ pairs, and $R$ is a set of $(B_r, G_{B_r})$ pairs. For each $B_u$, the algorithm processes node representation unlearning $\omega$ times, neighborhood reconstruction $\omega/2$ times. A high $\omega$ contributes to effective unlearning as it iterates $B_u$ $\omega$ times. However, it also means increasing computation time. After a full round of unlearning (a full pass over $U$), it checks the termination condition.

\begin{algorithm}[]
\caption{Node Representation Unlearning}
\small
\begin{algorithmic}[1]
    \Require $f$, $f_{\mathcal{E}}(\cdot)$, $B_{u}, G_{B_u}, B_r, G_{B_r}$  \\
    \textbf{Output} $f^\prime$
    \State $H_u = f_{\mathcal{E}}\left(B_{u}, G_{B_u} \right)$
    \State $H_r = f_{\mathcal{E}}\left(B_{r}, G_{B_r} \right)$
    \State $y_r = f\left(B_r, G_{B_r}\right)$
    \State $\ell_U = \mathcal{L}_U\left(H_u, H_r\right)$
    \State $\ell_{C} = \mathcal{L}_C\left(y_r, Y_{B_r}\right)$
    \State $f \leftarrow f - \eta\nabla\left(\beta\ell_{C} + \ell_{U}\right)$
    \State $f^\prime \leftarrow f$
    \State \textbf{return} $f^\prime$
\end{algorithmic}
\label{alg:node_representation_unlearning}
\end{algorithm}

\begin{algorithm}[]
\caption{Neighborhood Reconstruction}
\small
\begin{algorithmic}[1]
\Require $f_{\mathcal{E}}$, $\mathcal{N}$, $G_N$ \\
\textbf{Output} $f$
    \If{$\lvert N \rvert = 1$}
        \State $\mathcal{N}_1 = N \textrm{.POP\_FIRST()}$
        \State $G_{N_1} = G_N \textrm{.POP\_FIRST()}$
        \State $H_1 = f_{\mathcal{E}}\left(\mathcal{N}_1, G_{\mathcal{N}_1}\right)$
        \State $y_1 = f\left(\mathcal{N}_1, G_{\mathcal{N}_1}\right)$
        \State \textbf{return} $y_1$, $H_1$, $f$
    \Else
        \State $\mathcal{N}_1 = N\textrm{.POP\_FIRST()}$
        \State $G_{N_1} = G_N \textrm{.POP\_FIRST()}$
        \State $y_2, H_2, f = \textrm{\textsc{Neighborhood\_Reconstruction}}\left(f, N, G_N\right)$
        \State $H_1 = f_{\mathcal{E}}\left(\mathcal{N}_1, G_{\mathcal{N}_1}\right)$
        \State $\ell_N = \mathcal{L}_N\left(H_1, H_2\right)$
        \State $\ell_C = \mathcal{L}_C\left(y_2, Y_{\mathcal{N}_2} \right)$
        \State $f \leftarrow f - \eta\nabla\left(\ell_{C} + \ell_{R}\right)$
        \State $y_1 = f\left(\mathcal{N}_1, G_{\mathcal{N}_1}\right)$
        \State $H_1 = f_{\mathcal{E}}\left(\mathcal{N}_1, G_{\mathcal{N}_1}\right)$
    \EndIf \\
\textbf{return} $y_1$, $H_1$, $f$
\end{algorithmic}
\label{alg:neighborhood}
\end{algorithm}

Algorithm~\ref{alg:node_representation_unlearning} shows how node representation unlearning is conducted as a part of Node-CUL. It receives $B_u, G_{B_u}, B_r, G_{B_r}$, which are a batch of unlearning node, its subgraph, a batch of remaining node and its subgraph. %is a set of batches of unlearning nodes and their $k$-hop subgraphs. Similarly, $R$ refers to the set of batches of remaining nodes and their respective subgraphs. An unlearning round is defined as a single full pass over $U$. In each round, $B_u$ and $G_u$ are sampled from $U$. Then, for $\omega$ times, the algorithm samples $B_r$ and $G_r$ from $R$,
The algorithm obtains $H_u$ and $H_r$, which are the embeddings of $B_u$ and $B_r$ and computes contrastive loss (line 5).

Algorithm~\ref{alg:neighborhood} illustrates the neighborhood reconstruction algorithm. The algorithm receives $N$, a set of neighbors in ascending order. The first element is the first-hop neighbors, and the second element corresponds to the second-hop neighbors, and so on. It also receives $G_N$, a set of subgraphs of each element of $N$. The algorithm makes recursive calls, and in each call, the algorithm executes the line 9 and 10 that emit the first element of $N$ and $G_N$ using $.\textrm{POP\_FIRST()}$. To this end, neighbors that are closer to $B_u$ are popped out first. When only one element, which is the farthest neighbors, is left, the algorithm returns their predictions and embeddings. These are returned to the primitive function call, which holds predictions and embeddings of the one-step closer neighbors. Effectively, the farthest nodes are contrasted with one-step closer nodes, and closer nodes are subsequently contrasted and optimized.

While the entire algorithm requires multiple subgraphs to run, most of the subgraphs have overlapping nodes. Essentially, all subgraphs are subgraphs of a graph with nodes of $N^{K+1}_{B_u}$. Thus, once the graph is sampled, all subgraphs can be sampled from the graph. 

\section{Datasets and Hyperparameters}
\label{sec:implement_hyperparam}

\begin{table}[ht]
\centering
\small
    \begin{tabularx}{\linewidth}{*{5}{>{\centering\arraybackslash}X}}
        \toprule
         Dataset & Nodes & Edges & Features & Classes \\
         \midrule
         Cora-ML &  2708 & 10556 & 1433 &7 \\
         PubMed & 19717 & 88651 & 500 &3 \\
         Citeseer & 3327 & 9228 & 3703 &6 \\
         CS & 18333 & 163788 & 6805 & 15 \\
        \bottomrule
    \end{tabularx}
    \caption{Statistics of datasets}
    \label{tab:dataset}
\end{table}

\begin{table}[ht]
    \centering
    \small
    \setlength{\tabcolsep}{4pt}
    \small
    \begin{tabularx}{\linewidth}{l|*{4}{>{\centering\arraybackslash}X}}
        \toprule
        Model & Cora-ML & PubMed & Citeseer & CS \\
        \midrule
        GCN & 86.67 & 88.78 & 78.01 & 94.43 \\
        GAT & 87.78 & 88.78 & 77.71 & 93.89 \\
        GIN & 87.78 & 86.91 & 79.22 & 90.56 \\
        \bottomrule
    \end{tabularx}
    \caption{Performance comparison across different models and datasets}
    \label{tab:orig_model}
\end{table}

\begin{table}[]
    \centering
    \small
    \begin{tabularx}{\linewidth}{ll|*{3}{>{\centering\arraybackslash}X}}
        \toprule
        \textbf{Model} & \textbf{Dataset} & \textbf{Repeat} & \textbf{Batch Size} & \textbf{Learning Rate} \\
        \midrule
        \multirow{4}{*}{GCN} 
        & Cora & 2 & 128 & 0.005 \\
        & PubMed & 8 & 256 & 0.005\\
        & Citeseer & 2 & 128 & 0.005 \\
        & CS & 8 & 64 & 0.001 \\
        \midrule
        \multirow{4}{*}{GAT}
        & Cora & 4 & 128 & 0.005 \\
        & PubMed & 8 & 64 & 0.001 \\
        & Citeseer & 2 & 64 & 0.005 \\
        & CS & 8 & 64 & 0.001 \\
        \midrule
        \multirow{4}{*}{GIN}
        & Cora & 6 & 64 & 0.0005 \\
        & PubMed & 8 & 128 & 0.0001 \\
        & Citeseer & 6 & 256 & 0.0005 \\
        & CS & 8 & 128 & 0.0001\\
        \bottomrule
    \end{tabularx}
    \caption{Hyperparameter settings for different models and datasets}
    \label{tab:hyperparams}
\end{table}

Table~\ref{tab:dataset} shows the statistics of the datasets we used throughout the experiments.
We conduct a grid search over the hyperparameter space to find the best set of hyperparameters over different models and datasets. Table \ref{tab:orig_model} shows the performance and Table~\ref{tab:hyperparams} shows all hyperparameters for our experiments. 

For $\beta$, we used a fixed value of 8 throughout the experiments Similarly, we used $\gamma=1$ throughout the entire experiments.

%\section{Codes and Implementations}
%\label{sec:code}
%We provide our code at \textcolor{blue}{\href{https://anonymous.4open.science/r/Node-CUL-E30D/}{an anonymized git repository}}.

\section{Additional experiments}
\label{sec:appdix_exp}

\subsection{Effect of neighborhood reconstruction}

\begin{table}[!h]
    \centering
    \small
    \begin{tabularx}{\linewidth}{l|l|*{2}{>{\centering\arraybackslash}X}}
        \toprule
        Dataset & Metrics & \textbf{With} reconstruction & \textbf{Without} reconstruction \\
        \midrule
        \multirow{3}{*}{Cora-ML} 
        & Test accuracy & 87.65$\pm$1.42 & 85.09$\pm$0.71 \\
        & Unlearn accuracy & 85.03$\pm$0.57 & 83.17$\pm$0.57 \\
        & Unlearn score & 2.63 & 1.92 \\
        \midrule
        \multirow{3}{*}{PubMed}
        & Test accuracy & 89.09$\pm$0.07 & 85.86$\pm$0.46 \\
        & Unlearn accuracy & 86.83$\pm$ 0.42 & 84.40$\pm$0.18 \\
        & Unlearn score & 2.26 & 1.40 \\
        \midrule
        \multirow{3}{*}{Citeseer}
        & Test accuracy & 78.21$\pm$0.37 & 75.80$\pm$0.37 \\
        & Unlearn accuracy & 71.93$\pm$2.55 & 70.23$\pm$1.17 \\
        & Unlearn score & 6.28 & 5.56 \\
        \midrule
        \multirow{3}{*}{CS}
        & Test accuracy & 93.59$\pm$0.25 & 92.83$\pm$0.09 \\
        & Unlearn accuracy & 89.81$\pm$0.36 & 90.14$\pm$0.39 \\
        & Unlearn score & 3.78 & 2.68  \\
        \bottomrule
    \end{tabularx}
    \caption{Performance evaluation of unlearning the GCN model with and without neighborhood reconstruction.}
    \label{tab:neighborhood}
\end{table}

We conduct an ablation study to assess the impact of the neighborhood reconstruction. Table~\ref{tab:neighborhood} shows the test accuracy, unlearn accuracy, and unlearn scores of the GCN model unlearned with and without neighborhood reconstruction. The model was trained with 90\% of the Cora-ML dataset and unlearning 10\% of the training data.

The purpose of neighborhood reconstruction is to increase the model performance by eliminating the impact of unlearning nodes from the neighbors. It steers the model to disregard the embeddings of unlearning nodes when predicting the neighbors. Once the node representation unlearning is done, embeddings of the unlearning nodes are modified as they are pushed towards the decision boundary. However, if predictions of the neighbors are still influenced by the unlearning nodes, the effect of node representation unlearning can propagate to neighbors and reduces the prediction accuracy of them.

By introducing the neighborhood reconstruction, embeddings of the neighbors are less affected by the unlearning nodes. Accordingly, through the neighborhood reconstruction, embeddings of neighbors are less affected by the node representation unlearning, and the model can retain prediction performance of neighbors, which contributes to the utility of the model (test accuracy).

To verify this, we compare our Node-CUL framework with and without neighborhood reconstruction. If our claim is valid, the unlearned model with neighborhood reconstruction should exhibit higher test accuracy.

Table~\ref{tab:neighborhood} shows the impact of neighborhood reconstruction. For all datasets, having neighborhood reconstruction increased the model utility with slight loss in unlearn score. Although the unlearn score has seen a slight increase, this change is minor compared to the significant improvement in the model's performance.

%The purpose of neighborhood reconstruction is to (1) disassociate unlearning nodes from the neighbor nodes and (2) assist the model in making correct predictions for the neighbors. 
%Table~\ref{tab:neighborhood} shows clear evidence that neighborhood unlearning was able to achieve both purposes. The first purpose can be seen from the unlearn score. The model with neighborhood unlearning had a slightly higher unlearn score than the model  without. The neighborhood reconstruction has disjointed the embeddings of unlearning nodes from embeddings of the neighbors; hence, the model shows a higher unlearn score. On the other hand, the model unlearned without reconstruction has a smaller unlearn score, showing that unlearning nodes and the rest of the nodes have more similar embeddings. Table~\ref{tab:neighborhood_03} also supports this point. When unlearning a larger number of nodes (30\% of the training dataset), more neighbors are affected by the node representation  unlearning loss. Hence, without the neighborhood reconstruction, the model suffers significant utility loss.

%Simultaneously, the second purpose of neighborhood reconstruction can be observed from the test accuracy, as the test accuracy of models unlearned with the neighborhood reconstruction achieves better utility. In summary, the experiment demonstrates that the neighborhood reconstruction is functioning as we have designed, and it is essential to achieve better performance and more effective unlearning.

\end{document}